\documentclass[letterpaper, 10 pt, journal, twoside]{ieeetran}
\usepackage{amsmath,amsfonts}
\usepackage{algorithmic}
\usepackage{algorithm}
\usepackage{array}
\usepackage{textcomp}
\usepackage{stfloats}
\usepackage{url}
\usepackage{verbatim}
\usepackage{graphicx}
\usepackage{cite}

\usepackage{xcolor}
\usepackage{booktabs}
\usepackage{CJKutf8}
\usepackage{tabularx}
\usepackage{makecell}
\usepackage{subfigure}
\usepackage{float}
\usepackage{pifont}
\usepackage{multirow}

\begin{document}

\title{Observe Then Act: Asynchronous Active Vision-Action Model for Robotic Manipulation}

\author{Guokang Wang\textsuperscript{1}, Hang Li\textsuperscript{2}, Shuyuan Zhang\textsuperscript{3}, Di Guo\textsuperscript{2}, Yanhong Liu\textsuperscript{1,*}, and Huaping Liu\textsuperscript{4,*} 
\thanks{Manuscript received: October, 1, 2024; Revised December, 28, 2024; Accepted January, 27, 2025.}
}

\markboth{IEEE Robotics and Automation Letters. Preprint Version. Accepted January, 2025}
{Wang \MakeLowercase{\textit{et al.}}: Observe Then Act: Asynchronous Active Vision-Action Model for Robotic Manipulation} 


\maketitle

\begin{abstract}
In real-world scenarios, many robotic manipulation tasks are hindered by occlusions and limited fields of view, posing significant challenges for passive observation-based models that rely on fixed or wrist-mounted cameras.
In this paper, we investigate the problem of robotic manipulation under limited visual observation and  propose a task-driven asynchronous active vision-action model. Our model serially connects a camera Next-Best-View (NBV) policy with a gripper Next-Best-Pose (NBP) policy, and trains them in a sensor-motor coordination framework using few-shot reinforcement learning. This approach enables the agent to reposition a third-person camera to actively observe the environment based on the task goal, and subsequently determine the appropriate manipulation actions.
We trained and evaluated our model on 8 viewpoint-constrained tasks in RLBench. The results demonstrate that our model consistently outperforms baseline algorithms, showcasing its effectiveness in handling visual constraints in manipulation tasks. Our code and demonstration videos are available at \url{https://hymwgk.github.io/ota/}.
\end{abstract}


\section{Introduction}
\IEEEPARstart{A}{lthough}  reinforcement learning-based models for robotic manipulation can learn to infer actions directly from image observations through interaction with the environment, they still heavily rely on carefully positioned static third-person or wrist-mounted cameras\cite{james2022q,james2022coarse,grimes2023learning}. While these setups reduce sample variance during interaction\cite{grimes2023learning,cetin2022stabilizing}, they often struggle with providing sufficient visual data in occluded or limited-visibility scenarios due to unadjustable viewpoints or fixed camera-gripper configuration.

Alternatively, humans can adjust their head and eye positions according to task goals, which helps them obtain advantageous visual information for action decision-making \cite{land1999roles}. Inspired by this, recent active vision-action models allow agents to actively adjust their sensory viewpoints while interacting with the environment \cite{grimes2023learning, shang2024active, zaky2020active, lv2022sam}. These models, which mimic human capabilities, can be categorized as either synchronous or asynchronous \cite{srinivasan2010eye}. 

Synchronous models make decisions about visual and motor actions simultaneously based on current observations\cite{zaky2020active,lv2022sam,cheng2018reinforcement,uppal2024spin}. 
However, the need for simultaneous coordination of sensor-motor actions places high demands on the model's dynamic modeling capabilities and may limits its ability to handle tasks requiring large viewpoint shifts.
\begin{figure}
    \flushright
    \includegraphics[width=1.0\linewidth]{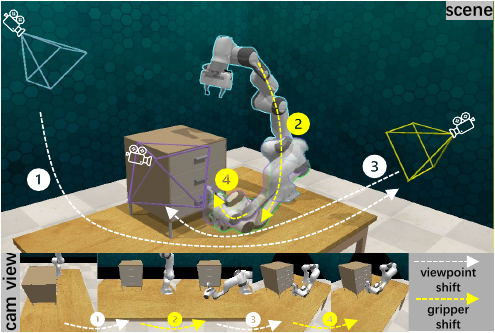}
    \caption{\textbf{Asynchronous active vision manipulation.} In the \textit{open\_drawer} task\cite{james2020rlbench}, the initial viewpoint may not effectively capture the drawer handle due to occlusion. Our model first predicts an optimal viewpoint for better observation of the handle and then determines the gripper action based on this updated view.}
    \vspace{-15pt}
    \label{fig:head_image}
\end{figure}
In contrast, asynchronous models decouple vision sensing and motor actions over time, allowing the agent to prioritize observation before interaction and then focus on inferring actions based on the selected viewpoint, thereby distributing sensor-motor coordination across the entire task episode. Although such models are commonly used for object grasping in occluded scenes\cite{morrison2019multi,chen2020transferable}, we believe that the asynchronous approach may also be suitable for sequential manipulation tasks.

\begin{figure*}[!ht]
    \centering
    \includegraphics[width=1.0\linewidth]{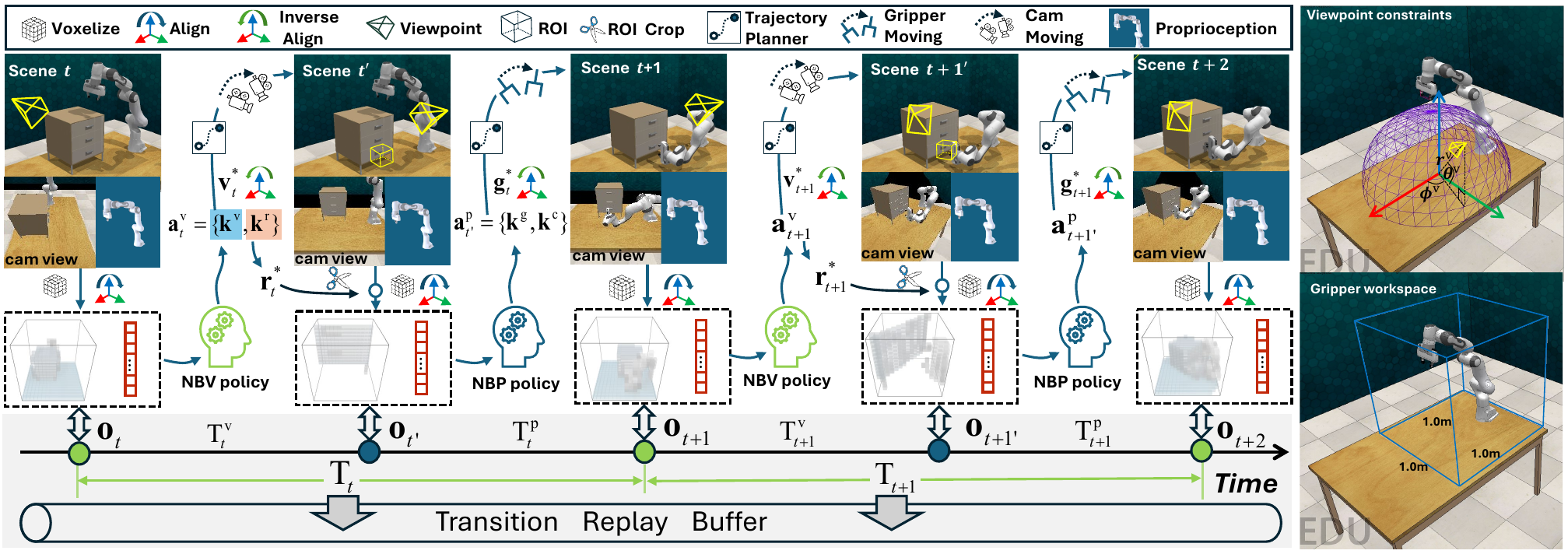}
    \caption{\textbf{Illustration of model pipeline and action spaces.} The NBV policy infers the 3D ROI position $\mathbf{r}^*$ based on the global scene observation and predict the optimal viewpoint $\mathbf{v}^*$ for the ROI observation  according to the given task goal. The NBP agent determines the gripper actions based on the ROI observation from $\mathbf{v}^*$.}
    \label{fig:framework}
    \vspace{-15pt}
\end{figure*}

In this paper, we introduce a task-driven asynchronous active vision-action model for robotic manipulation under limited observation (Fig.\ref{fig:head_image}). Our model employs a sequential dual-agent structure, with a Next-Best-View (NBV) agent to infer optimal viewpoints and a Next-Best-Pose (NBP) agent to determine gripper actions based on the observations from the NBV-inferred viewpoints (Fig.\ref{fig:framework}). By alternating between sensor and motor actions, the model actively perceives and interacts with the environment to achieve task objectives.
\IEEEpubidadjcol

We train our model with few-shot reinforcement learning\cite{james2022coarse} and introduce a viewpoint-aware demo augmentation method to maximize the utilization of limited demonstration trajectories. To address the expanded observation-action space caused by the active viewpoint, we employ a viewpoint-centric voxel alignment technique that rotates the scene voxel grid to align with the adjusted viewpoint. We also integrate a task-agnostic auxiliary reward to stabilize training, encouraging agents to interact effectively and select informative viewpoints for gripper action decisions. We evaluate our model on 8 RLBench\cite{james2020rlbench} tasks with limited viewpoints, demonstrating its effectiveness in viewpoint selection and action inference based on task requirements. 
The main contributions of this paper are as follows:
\begin{itemize}
    \item We propose a task-driven asynchronous active vision-action model that uses a novel dual-agent system to achieve effective robotic manipulation under limited observation.
    \item We introduce methodological innovations, including viewpoint-aware demo augmentation, viewpoint-centric voxel alignment, and a task-agnostic auxiliary reward strategy, to enhance model performance, learning efficiency, and training stability.
    \item Comprehensive experiments on 8 public RLBench tasks validate the effectiveness and robustness of our approach.
\end{itemize}

\section{Related Work}

\textbf{Active Vision for Manipulation.}
{Active vision \cite{aloimonos1988} aims to enable agents to actively observe their environment based on specific task goals, such as visual navigation\cite{uppal2024spin}, object recognition\cite{chaplot2021seal, fan2024evidential}, and 3D reconstruction \cite{zhao2022real, chen2024gennbv}.
In robotic manipulation, active vision is primarily applied to tasks such as object grasping \cite{zaky2020active, morrison2019multi, chen2020transferable, zhang2023affordance} and fruit harvesting \cite{burusa2022attention, burusa2024gradient}.
For more general manipulation tasks, Cheng et al. \cite{cheng2018reinforcement} trained an RL-based viewpoint policy to detect occluded objects and perform a pushing task. 
Lv et al. \cite{lv2022sam} trained a model-based active sensing-action model for flexible object manipulation. Chuang et al. \cite{chuang2023} used imitation learning to train an active viewpoint strategy that provides low-occlusion observations for peg insertion tasks. Shang et al. \cite{shang2024active} developed a self-supervised reward mechanism to jointly learn active visual sensing and motor skills for door opening and table wiping tasks. Dengler et al. \cite{dengler2024} trained an active viewpoint policy to learn how to push objects in confined spaces.
Our work is most similar to \cite{cheng2018reinforcement}, but unlike their model, we do not rely on a task-specific object detector. Furthermore, our active vision-action model uses an RL-based dual-agent asynchronous decision-making structure, allowing the agent to observe the environment first and then interact with it, similar to human behavior.}

\textbf{3D Voxel-based Manipulation.}
3D voxels are highly effective in representing spatial structures\cite{mescheder2019occupancy}, making them widely used in domains such as medical imaging\cite{you2022momentum,you2022simcvd} and autonomous driving \cite{tian2024occ3d, tong2023scene}.  Recently, they have also been adopted in robotic manipulation models for visual representation. For instance, James et al.\cite{james2022coarse} transformed original RGB-D images into 11-dimensional voxel grids and developed a 3DUnet-based reinforcement learning model to extract scene features and infer manipulation actions. Similarly, Shridhar et al.\cite{shridhar2023perceiver} used discrete 3D voxel images as input, employing the Perceiver network\cite{jaegle2021perceiver} as a backbone for end-to-end behavioral cloning, trained on demonstration trajectories. Leveraging the strong spatial structure representation and robust structural invariance of voxels under viewpoint changes—which aids in viewpoint learning—our model also uses voxelized scene images as input.

\begin{figure}
    \centering
    \includegraphics[width=1.0\linewidth]{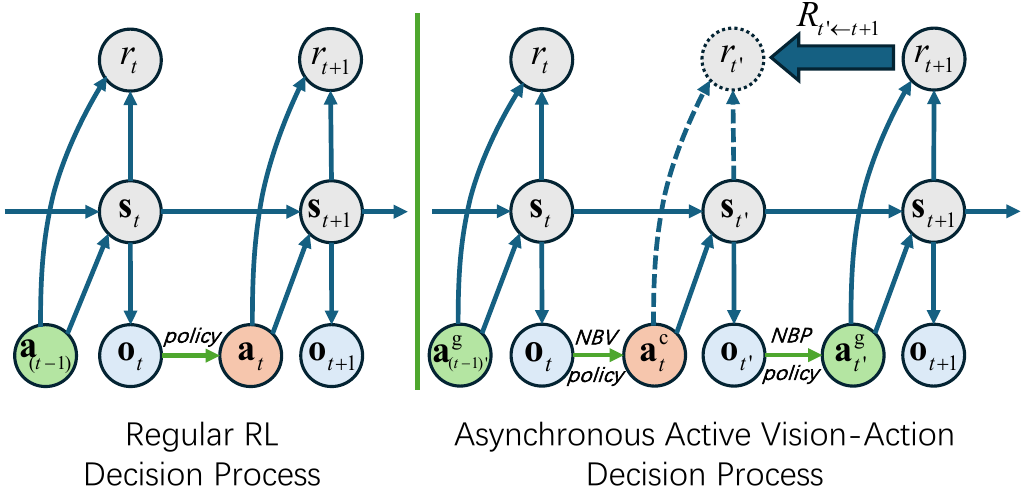}
    \caption{\textbf{Decision process comparison.} We divide single step interaction into NBV for viewpoint and NBP for action selection. Reward $r_\text{t'}$ is derived from $r_\text{t+1}$ after NBP interaction, enabling sensor-motor joint training through shared task rewards.}
    \vspace{-10pt}
    \label{fig:decision_process}
\end{figure}

\textbf{Attention-driven Robotic Manipulation.}
The attention mechanism, widely applied in fields such as object detection and natural language processing, contributes to robotic manipulation by focusing on the most task-relevant scene information from raw image inputs\cite{galassi2024attention}, improving both training efficiency and model performance. Gualtieri et al.~\cite{Marcus2018} generate grasp actions by applying attention to point cloud images to select noteworthy local regions and performing fine perception on these regions. ARM~\cite{james2022q} and C2FARM~\cite{james2022coarse} use attention mechanisms to focus on key visual input regions, with ARM selecting noteworthy regions from 2D inputs, while C2FARM predicts 3D regions of interest (ROI) from voxelized 3D images.  Liang et al.~\cite{liang2024visarl} pre-train a scene saliency attention encoder offline using human-annotated saliency maps and then train the reinforcement learning operation agent online based on the scene features output by this encoder. However, these methods largely rely on single or multiple fixed cameras from different viewpoints to capture images of the scene, lacking the capability to actively choose their viewpoints. {Our model incorporates the attention mechanism from C2FARM to select noteworthy local 3D regions from the scene but further enhances our model with the ability to predict the optimal observation angle for these regions, further mimicking human active visual perception capabilities.}

\section{Problem Formulation}
We divide one interaction time step $t\sim t+1$ into the viewpoint decision and transform process of the NBV policy $\pi_\text{v}$ from $t\sim t'$  and the gripper action decision and interaction process of the NBP policy $\pi_\text{p}$ from $t' \sim t+1$. 
{The asynchronous active vision manipulation problem can be formalized as a POMDP described by the tuple  $<\mathcal{S}, \mathcal{O}, \mathcal{A^\text{v}}, P, \mathcal{A^\text{p}}, P', r, \gamma>$, as shown in Fig.\ref{fig:decision_process}, where $\mathcal{S}$ is the environment states, $\mathcal{O}$ is the observation space, $\mathcal{A^\text{v}}, \mathcal{A^\text{p}}$ is the NBV and NBP action space, respectively. 
At time $t$, the NBV policy $\pi_\text{v}$ selects a viewpoint action $\mathbf{a}_t^\text{v} = \pi_\text{v}(\cdot|\mathbf{o}_t) \in \mathcal{A^\text{v}}$ based on the observation $\mathbf{o}_t \in \mathcal{O}$, resulting in a new observation $\mathbf{o}_{t'} \sim P(\cdot|\mathbf{s}_{t}, \mathbf{a}_t^\text{v})$. The NBP policy $\pi_\text{p}$ then selects a gripper action $\mathbf{a}_{t'}^\text{p} = \pi_\text{p}(\cdot|\mathbf{o}_{t'}) \in \mathcal{A^\text{p}}$, with the next scene observation $\mathbf{o}_{t+1}\sim P'(\cdot|\mathbf{s}_{t'}, \mathbf{a}_{t'}^\text{p})$ and receives a reward $r_{t+1}\in \mathbb{R}$. 
Our goal is to jointly learn $\pi^*_\text{v}$ and $\pi^*_\text{p}$ to maximize the expected sensor-motor reward $\mathbb E_{\pi_\text{v},\pi_\text{p}} \Big[\sum_{t=0}^\infty \gamma^t\cdot r(\mathbf s_t,\mathbf{a}^\text{v}_t,\mathbf{a}^\text{p}_{t'},\mathbf{s}_{t+1})  \Big]$.}

\textbf{Observation and Action Space.} The observation spaces $\mathcal{O}=\{\mathcal{O}^\text{i}, \mathcal{O}^\text{s}\}$, where $\mathbf{o}^\text{i} \in \mathcal{O}^\text{i}$ represents the 3D voxelized scene image from the  RGB-D data, and  $\mathbf{o}^\text{s} \in \mathcal{O}^\text{s}$ is the proprioceptive state. 
Specifically, the 3D voxel grid $\mathbf{o}^\text{i} \in \mathbb{R}^{n^2\times (3+M+1)}$ contains the voxel coordinates, $M$-dimensional features, and one occupancy flag for each voxel, where $n=16$ and $M=11$. {The proprioceptive state $\mathbf{o}^\text{s}=\{\mathbf{v}, \mathbf{g},d\}$ comprises the viewpoint spherical coordinates $\mathbf{v}=(r^\text{v}, \theta^\text{v}, \phi^\text{v})$, the gripper pose $\mathbf{g} \in \mathbb{R}^6$ and the gripper finger tips distance $d\in \mathbb{R}$.  
The camera’s movement is limited to the upper hemisphere at a fixed radius $r^\text{v}=1.2\text{m}$ above the table, illustrated in Fig.\ref{fig:framework}, with the optical axis directed at the sphere center. The NBV action $\mathbf{a}^\text{c}=\{\mathbf{k}^\text{v}, \mathbf{k}^\text{r}\}$, where $\mathbf{k}^\text{v} \in \mathbb{Z}^2$ is the discrete viewpoint index for $\theta^\text{v}\in [10^\circ, 60^\circ]$ and $\phi^\text{v} \in [-135^\circ, 135^\circ]$ with a resolution of $5^\circ$. $\mathbf{k}^\text{r} \in \mathbb{Z}^3$ denotes the position index of the 3D ROI that requires attention within the workspace. The $1\text{m}^3$ workspace is discretized into $16^3$ uniform grids, resulting in each dimension of $\mathbf{k}^\text{r}$ having an index range of $[0,15]$.
The NBP action $\mathbf{a}^\text{g}=\{\mathbf{k}^\text{g},\mathbf{k}^\text{c}\}$ consists of the discrete gripper pose index $\mathbf{k}^\text{g}\in \mathbb{Z}^6$ and desired gripper closure state $\mathbf{k}^\text{c}\in [0,1]$. The gripper position is constrained within the $0.3\text{m}^3$ ROI, which is further discretized into $16^3$ grids. The gripper’s rotation is represented by Euler angles on each axis, discretized in the range $[0^\circ,360^\circ]$ with a resolution of $5^\circ$. 
It is important to note that while both NBV and NBP agents share the same observation format, NBV observes the entire scene, whereas NBP focuses only on the 3D ROI image, achieving an attention effect~\cite{james2022coarse}. }
\section{Method}

\subsection{Model Structure}
{
Both the NBV and NBP policies are implemented using a DQN network with a 3D U-Net backbone\cite{cciccek20163d}, parameterized as $\Theta$ and $\Phi$, respectively.  The NBV agent infers the center position index $\mathbf{k}^\text{r}$ of the ROI in the scene and the optimal viewpoint index $\mathbf{k}^\text{v}$ for the ROI region based on the preprocessed aligned observation $\mathbf{o}_t$, as shown in Eq.\eqref{eq:nbv_q_function}. It then interacts with the environment and moves to move the viewpoint to the corresponding goal position $\mathbf{v}^*$.
\begin{equation}
\label{eq:nbv_q_function}
\mathbf{a}_t^\text{v} = \arg\underset{\mathbf{k}^\text{v}, \mathbf{k}^\text{r}}{\max} Q_\Theta(\mathbf{o}_t, \mathbf{k}^\text{v}, \mathbf{k}^\text{r})
\end{equation}

The depth image data captured at time $t'$ is cropped and voxelized, yielding the voxel grid observation $\mathbf{o}^\text{i}_{t'}$ for the ROI region under the new viewpoint. The NBP agent then infers the gripper pose index $\mathbf{k}^\text{g}$ and closure action $\mathbf{k}^\text{c}$ based on observation $\mathbf{o}_{t'}=\{\mathbf{o}^\text{i}_{t'},\mathbf{o}^\text{s}_{t'}\}$.

\begin{equation}
\label{nbp_q_function}
\mathbf a_{t'}^\text{p} = \arg\underset{\mathbf{k}^{\text{g}},\mathbf{k}^\text{c}}{\max}Q_\Phi(\mathbf o_{t'} ,\mathbf{k}^{\text{g}},\mathbf{k}^\text{c})
\end{equation}

Finally, the arm trajectory is planned using an off-the-shelf trajectory planner and executed through multiple control steps to move the gripper to the goal position $\mathbf{g}^*$, completing one sensor-motor interaction and receiving the reward $r_{t+1}$. The transition of the global agent denote as $\text{T}_t = \{\text{T}_t^\text{v}, \text{T}_t^\text{p}, r_{t+1}\}$ , where $\text{T}_t^\text{v} =(\mathbf{o}_t, \mathbf{a}_t^\text{v}, \mathbf{o}_{t'})$ is the local transition of the NBV agent, and $\text{T}_t^\text{p} = (\mathbf{o}_{t'}, \mathbf{a}_{t'}^\text{p}, \mathbf{o}_{t+1})$ is the NBP agent's local transition.  The explored transitions are stored in the replay buffer along with demonstration data during training.}

\subsection{Viewpoint-centric Voxel Alignment}\label{sec:alig}
\begin{figure}
    \centering
    \includegraphics[width=1.0\linewidth]{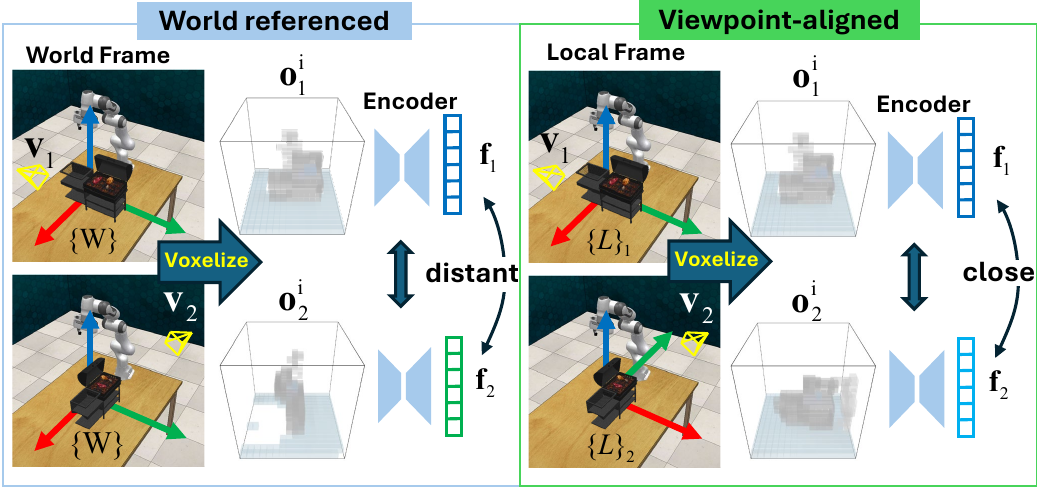}
    \caption{\textbf{Viewpoint-centric voxel alignment.} Due to the lack of spatial rotation invariance in 3D-CNNs, voxelizing the scene reference to world frame $\{\text{W}\}$ can cause similar observation from different viewpoints to appear distant in the feature space, increasing observation sample variance.}
    \label{fig:align}
    \vspace{-15pt}
\end{figure}

Images captured from a fixed viewpoint are primarily influenced by the scene state $\mathcal{S}$, following the mapping $\mathcal{S} \rightarrow \mathcal{O}^\text{i}$. While the active viewpoint setting introduces additional influence from viewpoint space $\mathcal{V}$, resulting in $\mathcal{S} \times \mathcal{V} \rightarrow \mathcal{O}^\text{i}$, which increases the variance of observation samples \cite{grimes2023learning}.    
To address this, we rotate the scene point cloud $\mathbf{o}^\text{pc}$ and world frame $\{\text{W}\}$ around the $z$-axis to align the world frame's $x$-axis with the viewpoint (Fig.\ref{fig:align}). This results in a local frame $\{\text{L}\}$ and aligned voxel grid image $\mathbf{o}^\text{i}$ after voxelization, as shown in Eq.\eqref{eq:obs-align}, where the rotation is defined as $^{\text L}{\text{T}}_{\text W}={\text{Rot}}^{-1}(z,\phi^\text{v})$.
\begin{equation}
\label{eq:obs-align}
\mathbf o^\text{i} = \mathbf{vox}(^{\text L}{\text{T}}_{\text W} \times \mathbf o^\text{pc})
\end{equation}

This alignment mitigates the variance fluctuations in observations mainly caused by viewpoint movements in the $\phi^\text{v}$ dimension. Since the inferred goal viewpoint $\mathbf{v}_\text{L}^*$, ROI center $\mathbf{r}_\text{L}^*$,and gripper pose $\mathbf{g}_\text{L}^*$are referenced to $\{\text{L}\}$, they need to be inverse-aligned to map the actions back to the world coordinate system for execution.

\vspace{-5pt}
\begin{equation}
\label{invers-align}
\mathbf{v},\mathbf{r},\mathbf{g} =^{\text{W}}{\text{T}}_{\text L}\times(\mathbf{v}_\text {L}^*,\mathbf{r}_\text{L}^*,\mathbf{g}_\text{L}^*)
\end{equation}

\subsection{Viewpoint-aware Demo Augmentation}\label{sec:aug}
\begin{figure}
    \centering
    \includegraphics[width=1.0\linewidth]{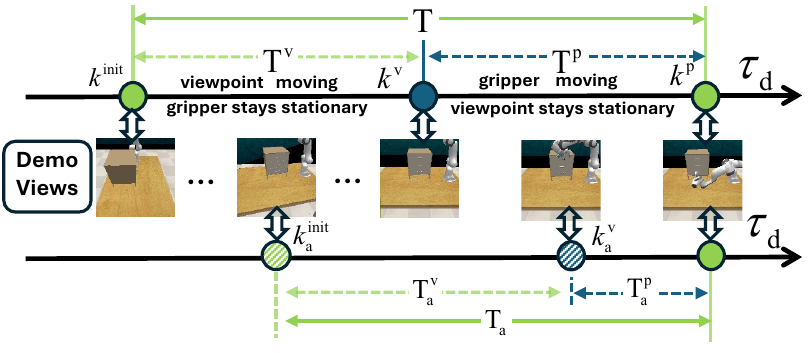}
    \caption{{\textbf{Demo trajectory augmentation.} The raw transition $\text{T}$ only contains observations and actions at key moments, while the additional augmentative transitions $\text{T}_\text{a}$ are constructed from intermediate observations and actions.}}
    \label{fig:demo_aug}
    \vspace{-15pt}
\end{figure}

{
During demonstration, the viewpoint and gripper movements are manually alternated asynchronously, staying stationary while the other is moving. These paired stationary viewpoints and gripper positions are called key actions, corresponding to the actions that need to be inferred by the NBV and NBP policies.
The paired data frame index of these key actions in one demo trajectory  $\tau_{\text{d}} = \{(\mathbf{o}_i, \mathbf{a}_i)\}^N$, are denoted as  $\big \{(k^{\text{v} },k^{\text{p}})\big\}^M\subset\mathbb R_+^2$ . Naively, we can use these key observation-action pairs to construct a prior transition sample set $\{\text{T}\}^M$  for active sensor-motor training, where $\text{T}$ defined as 
\vspace{-3pt}
\begin{equation}
\label{eq:raw_transition}
\text{T} = \{\text{T}^\text{v}, \text{T}^\text{p}, r\} = \{(\mathbf{o}_{k^{\text{init}}}, \mathbf{a}_{k^{\text{v}}}, \mathbf{o}_{k^{\text{v}}}), (\mathbf{o}_{k^{\text{v}}}, \mathbf{a}_{k^{\text{p}}}, \mathbf{o}_{k^{\text{p}}}), r\}.
\end{equation}

However, the number of these raw  transition samples is quite limited due to the small demonstration set. To fully utilize the demo trajectory data, we uniformly re-sample the initial observations of both $\text{T}^\text{v}$ and $\text{T}^\text{p}$ to construct additional transition samples for policy training, as shown in Fig.\ref{fig:demo_aug}. The resulting additional transitions is formalized as
\vspace{-2pt}
\begin{equation}
\label{eq:aug_transition}
\text{T}_\text{a} = \{\text{T}^\text{v}_\text{a}, \text{T}^\text{p}_\text{a}, r\} =\{(\mathbf{o}_{k^{\text{init}}_\text{a}}, \mathbf{a}_{k^{\text{v}}}, \mathbf{o}_{k^{\text{v}}_\text{a}}), (\mathbf{o}_{k^{\text{v}}_\text{a}}, \mathbf{a}_{k^{\text{p}}}, \mathbf{o}_{k^{\text{p}}}), r\}.
\end{equation}
}

Both $\text{T}$ and $\text{T}_\text{a}$ are ultimately stored in the replay buffer as early training experiences for the model.

\subsection{Task-agnostic Auxiliary Reward}\label{sec:aux}

{Relying solely on a centralized sparse reward to coordinate the collaboration between the NBV and NBP agents may result in only one agent being active while the other remains ``lazy", leading to a non-optimal equilibrium between the two agents \cite{sunehag2018value}.} To mitigate these issues and provide real-time guidance during the training process, we introduce two task-agnostic auxiliary rewards.

We introduce the $\textit{Interaction Encouragement Reward}$ $r_\text{i}$ to encourage the robot to interact with non-empty ROI areas containing scene entities. {If the NBP interaction does not fail due to trajectory planning errors or collisions, and the gripper's finger center point (GCP) is within the ROI after execution, we refer to this ROI as ``reachable".} The specific value assignment for $r_\text{i}$ is detailed as
\begin{equation}
\label{eq:ier_reward}
r_\text{i} =
\begin{cases}
0.02, & \text{ROI reachable and non-empty} \\
0.0, & \text{ROI reachable but empty} \\
-0.02, & \text{ROI unreachable}.
\end{cases}
\end{equation}

We introduce the $\textit{Entropy Reduction Reward}$ $ r_\text{e}$ to encourage NBV agent to select viewpoints that capture more determined visual information of the ROI. 
The reward $r_\text{e}$ is defined as $r_\text{e}=k \cdot \Delta E$, where $k$ is empirically set to 1.0 and $\Delta E$ is the total occupancy entropy change of the $N$ voxels inside the ROI after viewpoint adjustment. The occupancy entropy of the $i$-th voxel, $e_i$, is calculated using the Shannon entropy function \cite{burusa2024gradient}, shown in Eq.\eqref{eq:voxel_entropy}, where $p_\text{o}(i)$ represents its occupancy probability, and $q_\text{o}(i) = 1- p_\text{o}(i)$.

\begin{equation}
\label{eq:voxel_entropy}
\begin{split}
\Delta E &= E_t - E_{t'} = \frac{1}{N}\big(\textstyle\sum_{i=1}^N e_i^t - \textstyle\sum_{i=1}^N e_i^{t'}\big) \\
e_i &= -p_\text{o}(i)\log_2\big(p_\text{o}(i)\big) - q_\text{o}(i)\log_2\big(q_\text{o}(i)\big)
\end{split}
\end{equation}

The reward $r_\text{i}$ requires cooperation between the NBV and NBP policies, while $r_\text{e}$ is exclusive to the NBV policy. Therefore, the reward for updating the NBP policy is expressed as $r_\text{nbp} = r_\text{task} + r_\text{i}$, and the NBV reward is $r_\text{nbv} = r_\text{task} + r_\text{e} + r_\text{i}$, where $r_\text{task}\in[0,1]$ is determined by task completion. {These rewards are unrelated to specific task objectives, thus generalizing across tasks without the need for manual tuning.}

\section{Experiment Setup}
\begin{figure}
    \centering
    \includegraphics[width=1.0\linewidth]{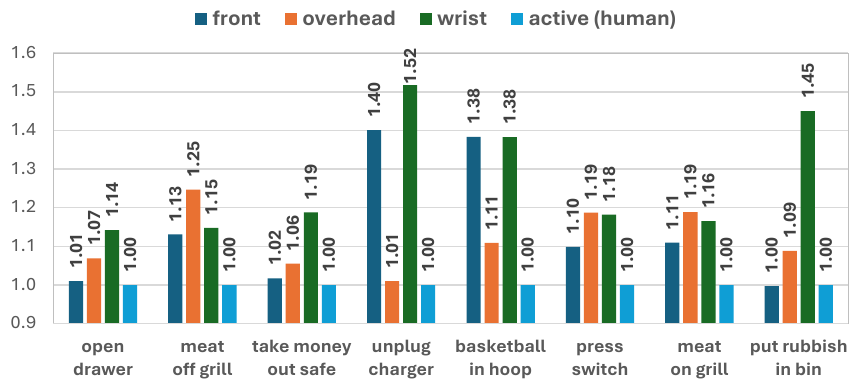}
    \caption{\textbf{Normalized viewpoint occlusion rate of selected tasks}. We consider human-demonstrated viewpoints to have the highest task prior knowledge and use them as the baseline for normalization.}
    \label{fig:NVOR}
    \vspace{-15pt}
\end{figure}

\textbf{Tasks and Initialization Settings.}
We trained and evaluated our model on RLBench \cite{james2020rlbench}, a widely used public benchmark for robotic manipulation. {The basic scene of RLBench is illustrated in Fig.\ref{fig:framework}, featuring several pre-set cameras, including fixed \textit{front} and \textit{overhead} views, as well as a \textit{wrist-mounted} camera attached to the gripper. At the start of each episode, the scene objects are randomly positioned within a legal region of $0.67\text{m}\times0.91\text{m}$ on the tabletop and randomly rotated around the $z$-axis of the world frame, which is perpendicular to the table. This ensures variation in scene initialization, guaranteeing differences in each training and testing episode.}
We selected 8 tasks (Table \ref{tab:Average Success Rate}) from RLBench following two criteria: 1) each task must include at least one RLBench default viewpoint with significant visual occlusion, and 2) no single fixed or wrist-mounted camera provides consistently low occlusion across the selected tasks. To evaluate the occlusion level of tasks and viewpoints, we define the Viewpoint Occlusion Rate (VOR) for a given viewpoint as the average occupancy entropy of the ROI where the gripper's goal actions occur. This is calculated based on observations from that viewpoint across $L = 100$ demonstration trajectories, as described in Eq.\eqref{eq:TOR}. Here, $E^{l,k}$, represents the ROI occupancy entropy at the $k$-th keyframe in trajectory $l$, after the viewpoint adjustment. Furthermore, we normalize the VOR of all viewpoints with reference to the VOR of the human-demonstrated viewpoint in the same task. This normalization facilitates intuitive comparisons across tasks and viewpoints. The results are shown in Fig.\ref{fig:NVOR}.
\begin{equation}
\label{eq:TOR}
\text{VOR}=\textstyle\frac{1}{L}\textstyle\sum_{l=1}^{L}\sum_{k=0}^{K} E^{l,k}
\end{equation}

\begin{figure}
    \centering
    \includegraphics[width=1.0\linewidth]{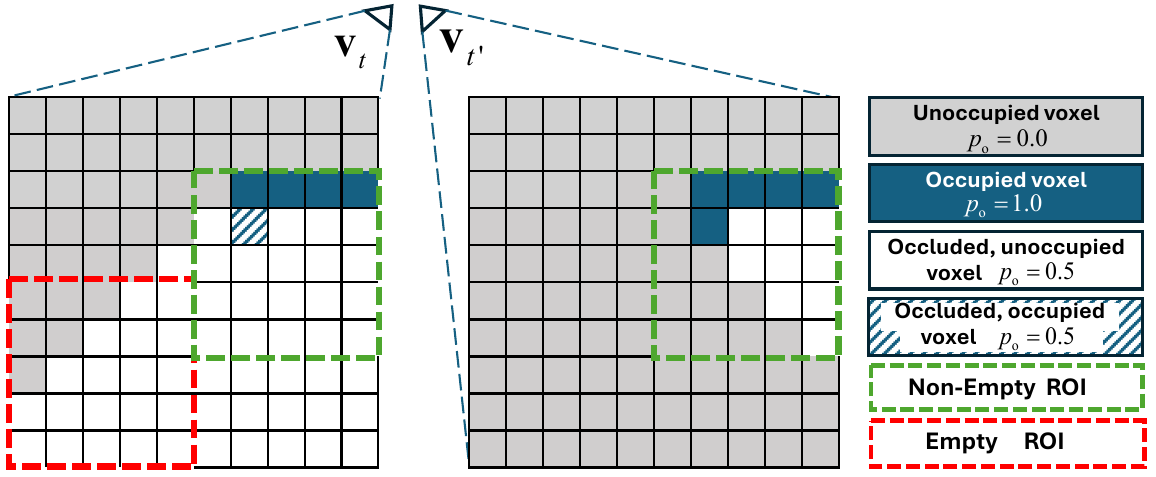}
    \caption{\textbf{Viewpoint adjustment and ROI selection in a 2D schematic scene.} {We define an ROI as non-empty only if it contains at least one voxel with an occupancy probability of $p_\text{o} = 1.0$; otherwise, it is considered empty.}}
    \label{fig:aux_task}
    \vspace{-15pt}
\end{figure}

\textbf{Baselines.}
We compared our model against three baseline categories: 1) visual manipulation models that utilize static third-person and wrist-mounted cameras, {2) active third-person camera-based models, and 3) Oracle baseline, which utilizes fully fused image from all default camera views in RLBench.
For the first category, we compared our model to C2FARM (C2F) \cite{james2022coarse} using static \textit{front}, \textit{overhead}, \textit{wrist} and \textit{front + wrist} (f+w) camera settings to evaluate the effectiveness of our active vision-action model in handling tasks with limited observability.}
For the second category, we evaluated our model against Behavior Cloning (BC) to test its ability to learn new sensor-motor skills beyond demonstration imitation. Additionally, we included a Random Viewpoint model \cite{grimes2023learning}, which shares the same architecture as ours but with the NBV policy network disabled (learning rate set to zero). This setup assesses the performance gain from active viewpoint adjustment versus random selection.
{For the Oracle baseline, we disabled the NBV viewpoint selection ability since full camera views were always accessible, but retained NBV's ROI inference capability.}

\textbf{Training Setting.}
We collected 30 demonstration trajectories for each task, for both ours and baseline methods. used them as prior knowledge for training. The Behavior Cloning model was trained using MSE loss, while the Random model used the same reward as our model but set the NBV learning rate to zero. All models were trained using Adam with a learning rate of $0.0005$ for 40,000 updates. Experiments were conducted on a machine with 4 NVIDIA RTX 4090 GPUs.

\section{Result}

\textbf{Main Performance.} \textbf{Main Performance.} We first analyzed the task Success Rate (SR) and Episode Length (EL) across all methods on 8 tasks to evaluate their overall performance. Table \ref{tab:Average Success Rate} shows that our model achieved the highest average success rate, outperforming the second-best single camera setting (C2F+front) by 26\%, especially in occluded tasks like \textit{meat\_on\_grill}, \textit{meat\_off\_grill} and \textit{unplug\_charger}. Similarly, Table \ref{tab: Average Episode Length} shows that our model had the shortest average episode length, outperforming the second-best single camera setting (C2F+front) by 28.2\% , demonstrating its high proficiency in the tasks.

We found that C2F with single camera setting performed well in low-occlusion tasks, such as \textit{basketball\_in\_hoop}, \textit{press\_switch}, \textit{put\_rubbish\_in\_bin}, and it occasionally matching or surpassing our method. However, their performance dropped sharply in occluded scenarios, leading to a higher performance variance. In contrast, our model remained robust in occluded tasks due to its active perception ability. { Leveraging complementary information from different viewpoints, C2F with a dual-camera setup (front + wrist) outperforms a single camera significantly but remains slightly below our single active camera setup, with both delivering comparable results. The Oracle baseline, benefiting from global scene perception, achieved the highest success rate and shortest episode length.}
In addition, although the BC model demonstrated some ability to adjust its viewpoint (as detailed below),  it struggled with limited training samples, achieving only a 14\% success rate. Meanwhile the random viewpoint policy performed the worst across all tasks, highlighting its inability to learn sensor-motor coordination.

\begin{table*}  
\footnotesize
\caption{\textbf{Task success rate}$\uparrow$ . Mean and std of 100 tests are reported (3 seeds).}
\label{tab:Average Success Rate}
\tabcolsep 5.5pt 
\begin{tabular}{l|cccccccc|c} 
\toprule
    Methods & \makecell{Basketball\\in Hoop}& \makecell{Meat \\off Grill}& \makecell{Meat\\on Grill} & \makecell{Open\\Drawer} & \makecell{Press\\Switch}  & \makecell{Put Rubbish \\in Bin
} & \makecell{Take Money\\out Safe} & \makecell{Unplug\\Charger}  & \makecell{On\\Average}\\\hline
    C2F (wrist)&{0.93}±0.02&0.01±0.01&	0.11±0.02&	0.44±0.07&	\textbf{0.90}±0.02	&0.00±0.00&	0.52±0.01	&0.30±0.02&	0.40±0.35 \\
    C2F (overhead)&0.87±0.02	&0.00±0.00&	0.28±0.05&	0.03±0.01&	0.89±0.03&	0.61±0.04	&0.24±0.03	&0.89±0.04	&0.48±0.36\\
    C2F (front)&0.26±0.03&	0.28±0.04&	0.68±0.04	&0.37±0.04&	0.86±0.06&	0.91±0.02	&0.66±0.06	&0.00±0.00	&0.50±0.30 \\
    {C2F (f+w)}&{\textbf{0.95}±0.03} &{0.32±0.04} &{0.65±0.04} &{\textbf{0.55}±0.06} & {0.84±0.04} & {0.92±0.02}&{\textbf{0.85}±0.04} & {0.73±0.01} & {0.72±0.23}\\
    Random (active) &0.00±0.00&  0.00±0.00&  0.00±0.00&  0.00±0.00& 0.00±0.00 & 0.29±0.02 & 0.00±0.00 &  0.00±0.00 & 0.04±0.09 \\
    BC (active)&0.00±0.00&	0.00±0.00&	0.02±0.01	&0.19±0.03	&0.51±0.02	&0.00±0.00	&0.38±0.03&	0.02±0.01&	0.14±0.19 \\
    Ours (active) &0.81±0.02&\textbf{0.56}±0.04&\textbf{0.73}±0.04&{0.50}±0.03&0.85±0.03&\textbf{0.96}±0.01	&{0.78}±0.02&\textbf{0.92}±0.01	&\textbf{0.76}±0.15 \\\hline
    {Oracle (full views)} &{0.95±0.02} &{ 0.93±0.03} & {0.91±0.05}&{0.82±0.03}&{0.92±0.01}& {0.98±0.01}&{0.97±0.02}&{0.96±0.04}& {0.93±0.05}\\
\bottomrule
\end{tabular}
\end{table*}

\begin{table*} 
\footnotesize
\caption{\textbf{Task episode length}$\downarrow$ . The maximum length of each task episode is limited to 10 steps.}
\label{tab: Average Episode Length}
\tabcolsep 4.5pt  
\begin{tabular}{l|cccccccc|c}  
\toprule
    Methods & \makecell{Basketball\\in Hoop}& \makecell{Meat \\off Grill}& \makecell{Meat\\on Grill} & \makecell{Open\\Drawer} & \makecell{Press\\Switch}  & \makecell{Put Rubbish \\in Bin
} & \makecell{Take Money\\out Safe} & \makecell{Unplug\\Charger}  & \makecell{On\\Average}\\\hline
    C2F (wrist) &{3.73}±0.18&	9.96±0.04	&9.27±0.20&	7.06±0.49&	\textbf{3.64}±0.23	&9.98±0.03&	6.39±0.11&	7.75±0.17&	7.22±2.39\\
    C2F (overhead)& 5.21±0.17	&10.00±0.01&	8.50±0.28&	9.78±0.07	&4.13±0.23	&6.52±0.21	&8.46±0.21&	3.93±0.22&	7.07±2.30\\
    C2F (front)&8.43±0.15&	8.66±0.18&	5.91±0.24	&8.04±0.20	&4.49±0.40	&\textbf{3.22}±0.13	&5.63±0.41	&9.99±0.02	&6.80±2.20\\
     {C2F (f+w)}&{\textbf{3.42}±0.21}&{7.51±0.26}&{6.14±0.25}&{\textbf{6.29}±0.35}&{4.81±0.34}&{3.13±0.14}&{5.60±0.17}&{5.22±0.11}& {5.27±1.48}\\
    Random (active) &10.00±0.00& 10.00±0.00& 10.00±0.00& 10.00±0.00& 10.00±0.00 & 7.76±0.27 & 10.00±0.00 & 10.00±0.00& 9.72±0.75   \\
    BC (active)& 10.00±0.00	&10.00±0.00	&9.93±0.05	&8.68±0.24&	6.74±0.28	&10.00±0.00	&7.64±0.27	&9.90±0.07	&9.11±1.22\\
    Ours (active) &5.35±0.08&	\textbf{6.43}±0.28&	\textbf{5.15}±0.26	&{6.65}±0.19	&3.89±0.28&	3.33±0.09	&\textbf{4.30}±0.17&	\textbf{3.91}±0.07&	\textbf{4.88}±1.16\\\hline
     {Oracle (full views)} &{3.41±0.13}&{4.56±0.16}& {4.33±0.38}&{4.32±0.32}&{3.49±0.24}& {3.08±0.01}&{3.78±0.04}&{3.12±0.29}& {3.76±0.58}\\
\bottomrule
\end{tabular}
\end{table*}

\begin{figure*}[htbp]
\centering
 \subfigure[]{
  \includegraphics[width=0.45\textwidth]{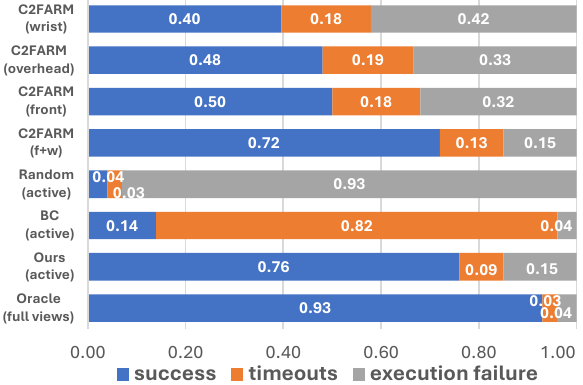}
  }%
  \hfill
  \subfigure[]{
  \includegraphics[width=0.48\textwidth]{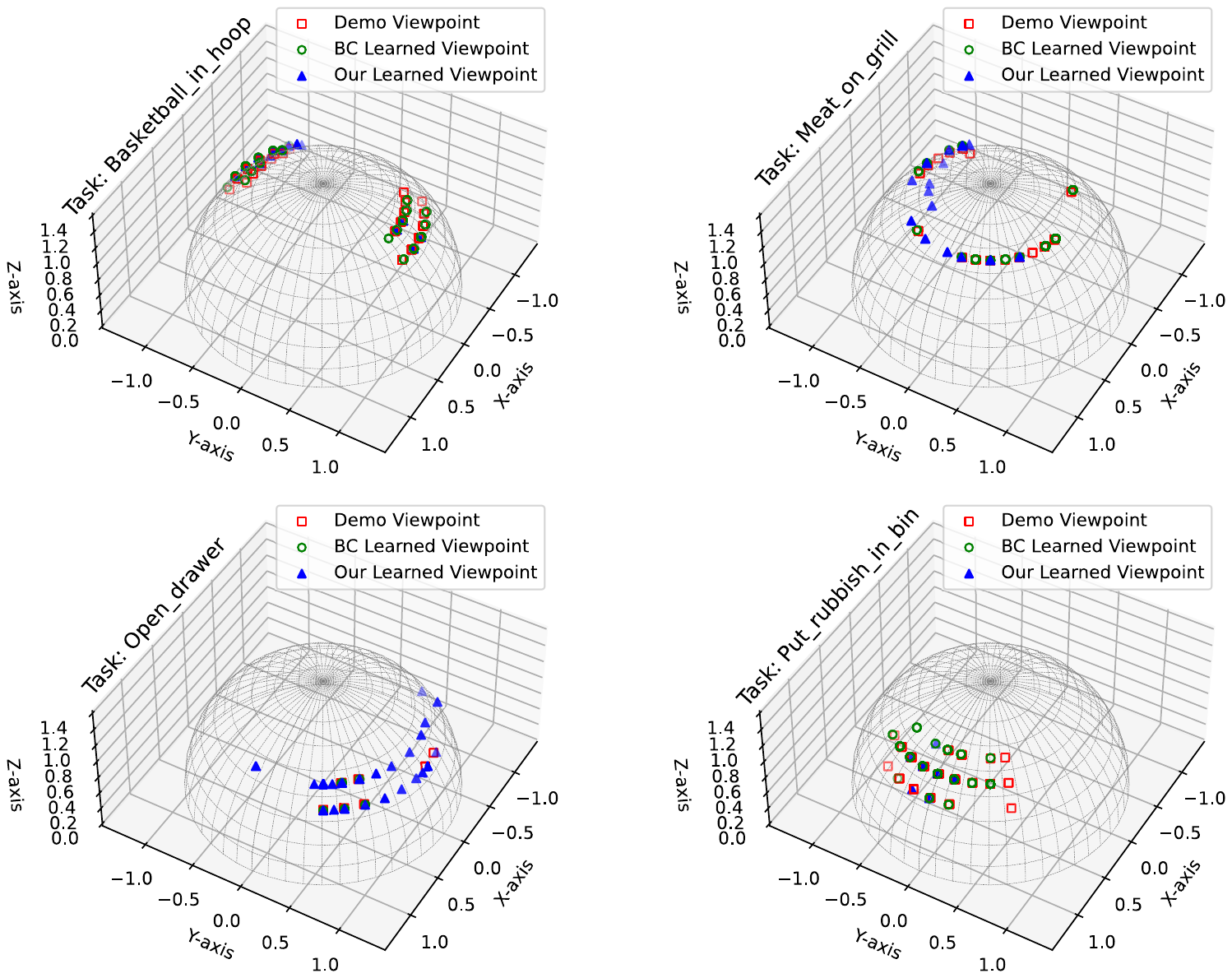}
  }
  \caption{\textbf{Result proportion and partial viewpoint distribution.} (a) Proportion of \textit{success}, \textit{timeouts}, and \textit{execution failures} across test episodes for different models. (b)Red squares indicate manually set viewpoints from 30 demo trajectories, while green circles and blue triangles represent the viewpoints learned by the behavior cloning and our model, respectively, from successful episodes across 100 tests.}
  \label{fig:rollout}
  \vspace{-10pt}
\end{figure*} 

\textbf{Active Perception Strategy Analysis.} {Active perception involves accurately identifying effective ROI regions in the scene and selecting optimal viewpoints to observe them, thereby supporting task completion. We propose Average Single Step Information Gain (ASSIG) and Average Initial Information Gain (AIIG) to quantitatively evaluate the model's ability to enhance ROI observation quality through viewpoint adjustments. ASSIG measures single-step information gain, while AIIG quantifies improvement relative to the initial viewpoint. They are defined as follows: 
\vspace{-3pt}
\begin{align}
\label{eq:ASSIG and AIIG definition}
\text{ASSIG}=\textstyle\frac{1}{L}\textstyle\sum_{l=1}^{L}\sum_{t=0}^{T}\Delta E^{l,t}\\
\text{AIIG}=\textstyle\frac{1}{L}\textstyle\sum_{l=1}^{L}\sum_{t=0}^{T}\Delta E^{l,t}_\text{init},
\end{align}

Where $L = 100$ testing episodes, $T$ is the episode length and $\Delta E$ are defined in Eq.\eqref{eq:voxel_entropy}. Fig.\ref{fig:rollout}(b) shows that both our model and the BC model demonstrate some learned viewpoint adjustment ability. Table.\ref{tab:Information Gain} further confirms this, with our model achieving up to 4.6\% and 7.0\% higher ASSIG and AIIG scores than the BC model.
However, as Fig.\ref{fig:rollout}(a) shows, the BC model has significantly more failures due to timeouts, indicating difficulty in accurately localizing ROI regions from limited demos. Additionally, our model learns more task-relevant viewpoints: some tasks (e.g., \textit{basketball\_in\_hoop}, \textit{put\_rubbish\_in\_bin}) show more focused distributions, while others (e.g., \textit{meat\_on\_grill}, \textit{open\_drawer}) exhibit greater generalization, reflecting further skill acquisition through interaction rather than mere imitation as seen with BC.}

\textbf{Environmental Modeling and Interactive Capabilities.} We evaluated the models' environment modeling and interaction capabilities by recording Success, Timeout (TO), and Execution Failure (EF) rates over 100 test episodes (Fig.\ref{fig:rollout}). The C2F model exhibited a high failure rate (up to 42\%) due to trajectory execution failures. Similarly, the Random model saw the majority of its test episodes end in failure because of execution errors. 
Although only 4\% of the BC model's episodes ended due to execution errors, 82\% failed due to timeouts. In contrast, our model showed a 13\% failure rate from trajectory errors and only 9\% from timeouts, indicating effective integration of low-level motion execution.

As shown in Table \ref{tab:Information Gain}, we also analyzed the average steps to First Interaction (FI) and Non-interactive steps (NI) per model. The BC model exhibited frequent non-interactive steps, correlating with its high timeout rate, suggesting difficulty learning precise ROI location and action skills from limited demonstrations. In contrast, our model initiated interactions more quickly, reducing non-interactive steps, and demonstrating superior accuracy in modeling environmental dynamics and structural understanding after training.

\begin{table} 
\footnotesize
\caption{Comparison of information gain and interaction capacity across different methods.}
\label{tab:Information Gain}
\tabcolsep 3.0pt  
\begin{tabular}{l|cc|cc} 
\toprule
    Methods & \makecell{ASSIG \(\uparrow\)\\(avg/max)} & \makecell{AIIG \(\uparrow\)\\(avg/max)}  & \makecell{First \(\downarrow\)\\Interaction} & \makecell{Non- \(\downarrow\)\\interaction}\\\hline
    C2F (wrist) & 0.0/0.0 & N/A &3.21±1.82&4.86±2.70\\
    C2F (overhead) &0.0/0.0 & 0.0/0.0 &2.97±1.70&4.41±2.77\\
    C2F (front) &0.0/0.0 & 0.0/0.0 &3.41±2.28&4.17±2.74\\
    C2F (f+w) &0.0/0.0 & N/A &1.91±0.55&2.63±1.32\\
    Random (active) &0.0/5.1\%&0.0/5.2\% &9.07±1.47&9.49±1.07\\
    BC (active) &0.7\%/2.3\%&5.6\%/18.1\% &6.31±2.46&7.54±2.15\\
    Ours (active) &\textbf{3.5}\%/\textbf{6.9}\%&\textbf{6.8}\%/\textbf{25.1}\%&\textbf{1.64}±0.61 &\textbf{2.15}±1.03\\\hline
    Oracle (full views) &0.0\%/0.0\%&N/A &1.58±1.36&1.86±1.56\\
\bottomrule
\end{tabular}
\vspace{-10pt}
\end{table}

\textbf{Ablation Study.} 
We first evaluated the impact of three strategies on manipulation performance: viewpoint-centric voxel alignment (Sec IV.B), viewpoint-aware demo augmentation (Sec IV.C), and task-agnostic auxiliary rewards (Sec IV.D). Table \ref{tab:ablation} shows that each strategy effectively improved success rates (SR), with voxel alignment and viewpoint-aware augmentation contributing the most, as success rates dropped by approximately 40\% without them and auxiliary rewards added around 6\% improvement. 

Beyond success rate improvements, all three strategies contributed to reductions in Episode Length (EL), Timeouts rate (TO), Execution Failures (EF), and the average number of Non-interactive steps (NI) per episode. This demonstrates their broader importance to the model's effectiveness.  Interestingly, while the voxel alignment strategy significantly boosts overall performance, we found it leads to a reduction in ASSIG score. We hypothesize that without viewpoint alignment, the NBV agent avoids potential issues with out-of-bounds actions and clipping, which may simplify the learning of viewpoint transformations. However, the overall performance still confirms that viewpoint alignment plays a crucial role in improving operational effectiveness.
Ablating all strategies led to a sharp decline, reducing the average success rate to 22\%, underscoring the synergistic effect of combining these strategies for optimal performance.

{We also evaluated our model with reduced demostrations. As shown in Table.{}, reducing the number of demos from 30 to 15 resulted in a 21\% drop in success rate. However, even with only 15 demos, the model slightly outperformed C2F with a single-camera setup trained on 30 demos, demonstrating its efficiency in leveraging limited demonstrations. Without any demos, the success rate dropped to 8\%, highlighting the significant challenges of learning active vision-action skills in a large observation-action space. Reducing the number of demos also increased episode lengths and execution failures, emphasizing the importance of providing sufficient, high-quality samples early in training. Such samples can help the model quickly learn environmental geometry and facilitate more efficient action exploration in subsequent stages. 

Interestingly, we observed that the model achieved meaningful ASSIG scores even without any demo data, slightly exceeding those obtained with 15 demos. We speculate that this is because when the global agent struggled to discover sparse task rewards, the local NBV agent may focused more on short-term entropy rewards $r_\text{e}$, leading to higher ASSIG scores. However, the AIIG scores, which require long-term planning, remained lower than those achieved with 15 demos.}

\begin{table} 
\footnotesize
\caption{Ablation over voxel alignment,demo augmentation and  task-agnostic auxiliary rewards.}
\label{tab:ablation}
\tabcolsep 3.0pt 
\begin{tabular}{ccc|cccccccc} 
\toprule
    \multicolumn{3}{c|}{Ablation Setting}   & \multicolumn{3}{c}{Result Proportion} & \multicolumn{3}{c}{Interactivity} & \multicolumn{2}{c}{Sensing}\\ \cmidrule(lr){1-3}\cmidrule(lr){4-6}\cmidrule(lr){7-9}\cmidrule(lr){10-11}
    align &aug &aux &SR$\uparrow$ &EF$\downarrow$&TO$\downarrow$&EL$\downarrow$&FI$\downarrow$&NI$\downarrow$& AIIG$\uparrow$ &ASSIG$\uparrow$\\\hline
    \ding{55}  &\checkmark &\checkmark &0.34&0.53&0.14 &7.61&4.33 &5.81&0.046&0.082\\
    \checkmark &\ding{55} &\checkmark &0.31&0.55&0.15 &7.99&4.49 &5.74&0.014&0.020\\
    \checkmark &\checkmark&\ding{55} &0.70&0.16&0.15 &5.52&1.85 &2.38&0.067&0.028\\
    \ding{55} &\ding{55} &\ding{55} &0.22&0.57&0.22 &8.46&5.74 &6.72&0.036&\textbf{0.083}\\
    \checkmark &\checkmark &\checkmark &\textbf{0.76}&\textbf{0.15}&\textbf{0.09}  &\textbf{4.88} &\textbf{1.64} &\textbf{2.15}&\textbf{0.068}&0.035\\  
\bottomrule
\end{tabular}
\vspace{-13pt}
\end{table}

\begin{table}
\footnotesize
\caption{{Ablation over different demo quantities.}}
\label{tab:demo_quantities}
\tabcolsep 5pt
\begin{tabular}{c|cccccccccc}
\toprule
    {Demo}   & \multicolumn{3}{c}{{Result Proportion}} & \multicolumn{3}{c}{{Interactivity}} & \multicolumn{2}{c}{{Sensing}}\\ \cmidrule(lr){2-4}\cmidrule(lr){5-7}\cmidrule(lr){8-9}
    {Num}  &{SR$\uparrow$} &{EF$\downarrow$}&{TO$\downarrow$}&{EL$\downarrow$}&{FI$\downarrow$}&{NI$\downarrow$}&{AIIG$\uparrow$} &{ASSIG$\uparrow$} \\\hline
    {0}   &{0.08}&{0.70}&{0.22} &{9.48}&{7.99} &{8.76}&{0.019}&{0.026}\\
    {15}  &{0.51}&{0.36}&{0.13} &{6.53}&{3.23} &{4.38}&{0.046}&{0.020}\\
    {30}  &{\textbf{0.76}}&{\textbf{0.15}}&{\textbf{0.09}}&{\textbf{4.88}}&{\textbf{1.64}} &{\textbf{2.15}}&{\textbf{0.068}}&{\textbf{0.035}}\\ 
\bottomrule
\end{tabular}
\vspace{-13pt}
\end{table}

\section{Conclusion and Future Work}
In this paper, we investigate the problem of robotic manipulation under limited observation and present a task-driven asynchronous active vision-action model.  Our model first infers the optimal viewpoint to actively observe the scene, then selects the manipulation action from the updated observation.  Our model was trained and evaluated on eight RLBench tasks with varying levels of occlusion, demonstrating that it significantly outperforms passive observation-based models by actively refining viewpoints and learning sensor-motor coordination through interaction with the environment. {However, one limitation is its reliance on a small number of demonstration samples to provide early training signals. Future work could explore strategies for zero-shot training and extend the approach to longer-horizon tasks.}



\end{document}